\documentclass[letterpaper]{article} 
\usepackage{aaai2026}  

\usepackage{times}  
\usepackage{helvet}  
\usepackage{courier}  
\usepackage[hyphens]{url}  
\usepackage{graphicx} 
\usepackage{natbib}  
\usepackage{caption} 
\usepackage{nameref}

\usepackage{amsmath}
\usepackage{multirow}

\usepackage{algorithm}
\usepackage{algpseudocode}
\usepackage{newfloat}
\usepackage{listings}
\DeclareCaptionStyle{ruled}{labelfont=normalfont,labelsep=colon,strut=off} 
\floatstyle{ruled}
\newfloat{listing}{tb}{lst}{}
\floatname{listing}{Listing}
\usepackage{booktabs}        
\usepackage{tabularx}        
\usepackage[table]{xcolor}   
\usepackage{graphicx}        
\usepackage{xcolor}
\usepackage{pifont}          
\usepackage{colortbl}        
\usepackage{array}           
\usepackage{caption}         

\definecolor{ForestGreen}{RGB}{34,139,34}
\definecolor{OrangeRed}{RGB}{255,69,0}

\usepackage{tabularx} 
\newcolumntype{Y}{>{\centering\arraybackslash}X}

\usepackage{tikz}
\usetikzlibrary{shapes.geometric, arrows.meta, positioning}

\tikzstyle{block} = [rectangle, draw, text centered, minimum height=2em, minimum width=5em]
\tikzstyle{arrow} = [->, thick]

\usepackage{booktabs}        
\usepackage{array}           
\usepackage{colortbl}        
\usepackage[table]{xcolor}   
\usepackage{graphicx}        
\usepackage{multirow}        
\usepackage{caption}         
\usepackage{amsmath}         
\usepackage{amssymb}         
\usepackage{pifont}          

\newcommand{\demph}[1]{\textcolor{gray}{#1}}

\usepackage{graphicx}
\usepackage{subcaption}
\usepackage[caption=false,font=footnotesize]{subfig}

\usepackage{enumitem}

\definecolor{ThemeColor}{HTML}{F0F8FF} 

\newcommand{\benchmarkname}{AURA}
\newcommand{\evalmetricname}{\texttt{AuraScore}}
\title{\benchmarkname{}: A Fine-Grained Benchmark and Decomposed Metric for Audio-Visual Reasoning}
\author{
    Siminfar Samakoush Galougah\textsuperscript{\rm 1},
    Rishie Raj\textsuperscript{\rm 1},
    Sanjoy Chowdhury\textsuperscript{\rm 1},
    Sayan Nag\textsuperscript{\rm 2},
    Ramani Duraiswami\textsuperscript{\rm 1}
}
\affiliations{
    \textsuperscript{\rm 1}University of Maryland, College Park \quad
    \textsuperscript{\rm 2}University of Toronto \\
    \{simin95, rraj27, sanjoyc, ramanid\}@umd.edu, sayan.nag@mail.utoronto.ca \\
    \texttt{https://rishieraj.github.io/aura-project-page/}
}

\begin{document}

\twocolumn[{
\renewcommand\twocolumn[1][]{#1}
\maketitle
\begin{center}
    \centering
    \captionsetup{type=figure}
    \includegraphics[width=\textwidth]{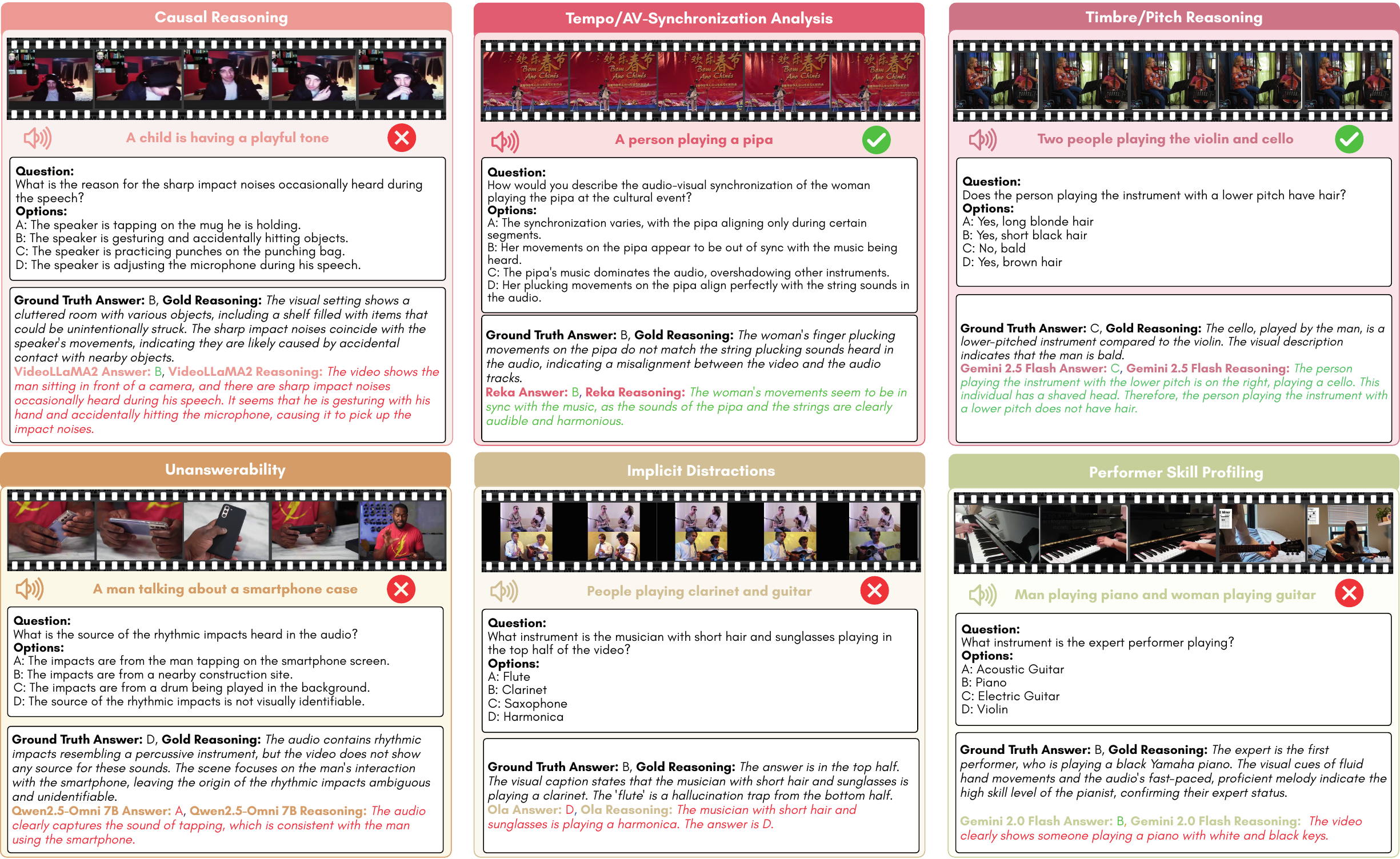}
    \captionof{figure}{\textbf{Introducing \benchmarkname{}}. We present \benchmarkname{}, the first question-answering (QA) benchmark designed to evaluate state-of-the-art Audio-Visual Large Language Models (AV-LLMs) and Omni-Modal Language Models (OLMs) on fine-grained cognitive tasks, including \textbf{\textit{Cross-Modal Causal Reasoning}}, \textbf{\textit{Timbre/Pitch Reasoning}}, \textbf{\textit{Tempo/AV Synchronization Analysis}}, \textbf{\textit{Unanswerability}}, \strut \textbf{\textit{Implicit Distractions}}, and \textbf{\textit{Performer Skill Profiling}}.}
    \label{fig:six_pngs}
\end{center}
}]

\begin{abstract}
Current audio-visual (AV) benchmarks focus on final answer accuracy, overlooking the underlying reasoning process. This makes it difficult to distinguish genuine comprehension from correct answers derived through flawed reasoning or hallucinations. To address this, we introduce \textit{\benchmarkname{} (Audio-visual Understanding and Reasoning Assessment)}, a benchmark for evaluating the cross-modal reasoning capabilities of Audio-Visual Large Language Models (AV-LLMs) and Omni-modal Language Models (OLMs). AURA includes questions across six challenging cognitive domains, such as causality, timbre and pitch, tempo and AV synchronization, unanswerability, implicit distractions, and skill profiling, explicitly designed to be unanswerable from a single modality. This forces models to construct a valid logical path grounded in both audio and video, setting AURA apart from AV datasets that allow uni-modal shortcuts. To assess reasoning traces, we propose a novel metric, \evalmetricname{}, which addresses the lack of robust tools for evaluating reasoning fidelity. It decomposes reasoning into two aspects: \textit{(i)} Factual Consistency - whether reasoning is grounded in perceptual evidence, and \textit{(ii)} Core Inference - the logical validity of each reasoning step. Evaluations of SOTA models on AURA reveal a critical reasoning gap: although models achieve high accuracy (up to 92\% on some tasks), their Factual Consistency and Core Inference scores fall below 45\%. This discrepancy highlights that models often arrive at correct answers through flawed logic, underscoring the need for our benchmark and paving the way for more robust multimodal evaluation.
\end{abstract}

\begin{table*}[t]
\centering
\small
\renewcommand{\arraystretch}{1.0}
\resizebox{\textwidth}{!}{
\begin{tabular}{lccccccccccc}
\toprule
\multirow{3}{*}{\textbf{Dataset}} & \multirow{3}{*}{\textbf{Domain}} & \multirow{3}{*}{\textbf{QA Type}} & \multirow{3}{*}{\textbf{QA Generation}} & \multirow{3}{*}{\textbf{Eval. Type}} & \multirow{3}{*}{\textbf{CMD}} & \multirow{3}{*}{\textbf{RTE}} & \multicolumn{2}{c}{\textbf{FGVA}} & \multicolumn{3}{c}{\textbf{FGAA}} \\
\cmidrule(lr){8-9} \cmidrule(lr){10-12}
 &  &  &  &  &  &  & DTR & SKL & TMP & PTH & TBR \\
\midrule
AVQA~\cite{avqa} & Daily-life & MCQ & Manual & Final Acc. & \textcolor{ForestGreen}{\ding{51}} & \textcolor{OrangeRed}{\ding{55}} & \textcolor{OrangeRed}{\ding{55}} & \textcolor{OrangeRed}{\ding{55}} & \textcolor{OrangeRed}{\ding{55}} & \textcolor{OrangeRed}{\ding{55}} & \textcolor{OrangeRed}{\ding{55}} \\
AVInstruct~\cite{avinstruct} & General & Open & Auto & LLM-based Final Acc. & \textcolor{ForestGreen}{\ding{51}} & \textcolor{OrangeRed}{\ding{55}} & \textcolor{OrangeRed}{\ding{55}} & \textcolor{OrangeRed}{\ding{55}} & \textcolor{OrangeRed}{\ding{55}} & \textcolor{OrangeRed}{\ding{55}} & \textcolor{OrangeRed}{\ding{55}} \\
Music-AVQA~\cite{r:47} & Music & Open & Manual & Word Match & \textcolor{ForestGreen}{\ding{51}} & \textcolor{OrangeRed}{\ding{55}} & \textcolor{OrangeRed}{\ding{55}} & \textcolor{OrangeRed}{\ding{55}} & \textcolor{OrangeRed}{\ding{55}} & \textcolor{OrangeRed}{\ding{55}} & \textcolor{OrangeRed}{\ding{55}} \\
VGGSound~\cite{vggsound} & General & Classification & Auto & Final Acc. / mAP & \textcolor{ForestGreen}{\ding{51}} & \textcolor{OrangeRed}{\ding{55}} & \textcolor{OrangeRed}{\ding{55}} & \textcolor{OrangeRed}{\ding{55}} & \textcolor{OrangeRed}{\ding{55}} & \textcolor{OrangeRed}{\ding{55}} & \textcolor{OrangeRed}{\ding{55}} \\
Daily-Omni~\cite{r:03}  & Daily-life & MCQ & Semi-Auto & Final Acc. & \textcolor{ForestGreen}{\ding{51}} & \textcolor{OrangeRed}{\ding{55}} & \textcolor{OrangeRed}{\ding{55}} & \textcolor{OrangeRed}{\ding{55}} & \textcolor{OrangeRed}{\ding{55}} & \textcolor{OrangeRed}{\ding{55}} & \textcolor{OrangeRed}{\ding{55}} \\
WorldSense~\cite{r:39}  & General & MCQ & Manual & Final Acc. & \textcolor{ForestGreen}{\ding{51}} & \textcolor{OrangeRed}{\ding{55}} & \textcolor{OrangeRed}{\ding{55}} & \textcolor{OrangeRed}{\ding{55}} & \textcolor{OrangeRed}{\ding{55}} & \textcolor{OrangeRed}{\ding{55}} & \textcolor{OrangeRed}{\ding{55}} \\
\midrule
VideoVista~\cite{r:01}  & General & MCQ & Auto & Final Acc. & \textcolor{OrangeRed}{\ding{55}} & \textcolor{OrangeRed}{\ding{55}} & \textcolor{OrangeRed}{\ding{55}} & \textcolor{OrangeRed}{\ding{55}} & \textcolor{OrangeRed}{\ding{55}} & \textcolor{OrangeRed}{\ding{55}} & \textcolor{OrangeRed}{\ding{55}} \\
OmniEval~\cite{r:02} & General & MCQ \& Open & Semi-Auto & Final Acc. \& LLM & \textcolor{ForestGreen}{\ding{51}} & \textcolor{OrangeRed}{\ding{55}} & \textcolor{OrangeRed}{\ding{55}} & \textcolor{OrangeRed}{\ding{55}} & \textcolor{OrangeRed}{\ding{55}} & \textcolor{OrangeRed}{\ding{55}} & \textcolor{OrangeRed}{\ding{55}} \\
\midrule
CG-Bench~\cite{r:04} & Long-form & MCQ \& Open & Manual & Final Acc. \& Clue & \textcolor{OrangeRed}{\ding{55}} & \textcolor{OrangeRed}{\ding{55}} & \textcolor{OrangeRed}{\ding{55}} & \textcolor{OrangeRed}{\ding{55}} & \textcolor{OrangeRed}{\ding{55}} & \textcolor{OrangeRed}{\ding{55}} & \textcolor{OrangeRed}{\ding{55}} \\
\textbf{\benchmarkname{} (Ours)} & \textbf{General \& Music} & \textbf{MCQ} & \textbf{Auto} & \textbf{\evalmetricname{}} & \textbf{\textcolor{ForestGreen}{\ding{51}}} & \textbf{\textcolor{ForestGreen}{\ding{51}}} & \textbf{\textcolor{ForestGreen}{\ding{51}}} & \textbf{\textcolor{ForestGreen}{\ding{51}}} & \textbf{\textcolor{ForestGreen}{\ding{51}}} & \textbf{\textcolor{ForestGreen}{\ding{51}}} & \textbf{\textcolor{ForestGreen}{\ding{51}}} \\
\bottomrule
\end{tabular}
}
\caption{\textbf{Comparison with prior Audio-Visual benchmarks.} This table highlights each benchmark's focus on mandatory cross-modal reasoning, reasoning trace evaluation, and the assessment of fine-grained visual and audio attributes. CMD: Cross-Modal Dependency, RTE: Reasoning Trace Evaluation, FGVA: Fine-Grained Visual Attributes (DTR: Distractions, SKL: Skill). FGAA: Fine-Grained Audio Attributes (TMP: Tempo/Synchronization, PTH: Pitch, TBR: Timbre).}

\label{tab:dataset_comparison}
\end{table*}

\section{Introduction} \label{intro}

The rapid evolution of multi-modal research, particularly in the audio-visual domain, has produced AV-LLMs \cite{r:10,r:18,r:19,r:20} and OLMs \cite{r:05, r:06, r:07, r:08, r:09} with remarkable capabilities. The ultimate goal is to achieve human-like reasoning, yet the methods used to evaluate progress are lagging significantly. Initial benchmarks like \cite{r:39} used a simple multiple-choice accuracy on real-world audio-visual questions. \cite{r:01} used the same MCQ accuracy framework for evaluation while differentiating the questions based on \textit{understanding} and \textit{reasoning}. Subsequently, \cite{r:03} was the first benchmark to filter out questions that were answerable using text-only and showed the lack of cross-modal temporal awareness in state-of-the-art MLLMs. Progress was also made in coupling answer accuracy with grounding in \cite{r:04, r:02}, where they utilized clue-based grounding and incorporated audio-visual-temporal grounding tasks, respectively. 

However, despite these advancements, none of the previous benchmarks have gone beyond simply evaluating the output of MLLMs. This approach is critically insufficient as it cannot distinguish between genuine comprehension and answers that are merely sophisticated hallucinations or the product of flawed logic. A classic example can be seen in Figure \ref{fig:six_pngs}, where under the \textit{Causal Reasoning} task, one of the evaluation models gets the answer \textit{"The speaker is gesturing and accidentally hitting objects."} correct, but the reasoning is wrong. This means that while the model might get the "cause of the sound" right, it hallucinates on the \textit{reason} behind the cause of that sound. Another exciting example can be seen in the \textit{Performer Skill Profiling} section of Figure \ref{fig:six_pngs}. Here, the evaluation model gets the answer correct on the instrument being played by the \textit{expert performer}, but when asked to share the reasoning behind its choice, the model responded with \textit{"The video clearly shows someone playing a piano with white and black keys."}. It can be seen that the model makes no assessment about any skill-based attribute of the performer, which leads us to believe the model might just be making a random guess.

Based on such extensive evaluations, we have identified two critical shortcomings in previous audio-visual benchmarks: \textit{(1)} They are not designed to assess advanced audio and visual attributes such as \textit{tempo}, \textit{pitch}, \textit{skill}, etc., and \textit{(2)} None of them evaluate the reasoning behind a chosen answer by the models. To address this evaluation gap, we introduce \textbf{\benchmarkname{}} (Audio-visual Understanding and Reasoning Assessment), which pioneers a new paradigm, focused on evaluating the reasoning traces for a model's response across 6 diverse and unique cross-modal tasks that focus on advanced, fine-grained audio-visual attributes.

To ensure robust evaluation of the reasoning traces, we propose \textbf{\evalmetricname{}}, a novel metric that decomposes a model's thought process into two fundamental components: \textit{(i)} \textbf{Factual Consistency}, which assesses whether the reasoning is grounded in perceptual evidence, and \textit{(ii)} \textbf{Core Inference}, which evaluates the logical validity of each step. This combined framework of challenging tasks and a diagnostic metric provides the first comprehensive tool to move beyond surface-level accuracy and truly scrutinize the faithfulness of a model's reasoning ability. We evaluated several state-of-the-art AV-LLMs and OLMs such as VideoLLama2~\cite{r:10}, Gemini 2.5~\cite{r:29}, and Reka~\cite{r:17} on our benchmark (results as shown in Table \ref{omnimodalsCmp}). These were our key findings:

\begin{itemize}
    \item There is a huge gap between answer accuracy scores and reasoning inference scores. While the best performing model (Qwen2.5-Omni 7B~\cite{r:07} for most tasks) usually scores quite high in accuracy ($\sim 80-90\%$), its reasoning scores, particularly reasoning inference, are pretty low ($\sim 20-40\%$). This supports our hypothesis that it is necessary to assess models beyond simple accuracy.
    \item Performance has been varied across all six tasks. While evaluation on most tasks has revealed high accuracy and low reasoning score, some tasks assess advanced audio attributes such as \textit{tempo/av-synchronization} and \textit{pitch/timbre}, where models have performed poorly across all metrics. This shows the importance of evaluating models across a wide range of tasks that assess fine-grained modal features. 
\end{itemize}

 While many models achieve high final-answer accuracy, it was observed that their performance plummets when evaluated on Factual Consistency and Core Inference, exposing a fundamental weakness in their cognitive process (as shown in Table \ref{tab:dataset_comparison}). This discrepancy proves that current models often rely on flawed logic rather than genuine comprehension. By systematically exposing these weaknesses, our work provides a clear and actionable roadmap for the community to develop more robust and faithful AVLLMs. Our primary contributions are as follows:

\begin{itemize}
    \item \textbf{A Benchmark for Advanced Cognitive Audio-Visual Reasoning.} We introduce AURA, a benchmark with reasoning tasks designed for unique, challenging, and underexplored audio-visual domains. These tasks, including performer skill profiling, tempo and audio-visual synchronization, and timbre recognition, probe subtle attributes and correlations that demand a deep level of reasoning.
    \item \textbf{A Novel Decomposed Evaluation Metric.} We propose \evalmetricname{}, the first metric to evaluate the reasoning trace of AV models by decomposing it into two aspects: \textit{(i)} Factual Consistency, ensuring the reasoning is grounded in perceptual evidence, and \textit{(ii)} Core Inference, assessing the logical validity of the thought process. This provides a fine-grained analysis that goes far beyond simple answer accuracy.
    \item \textbf{An Automated QA Generation Pipeline.} We develop a modular and scalable pipeline that automatically generates complex QA pairs for our unique domains. It uses fine-grained video and audio captions to construct challenging questions, and its design allows for easy adaptation and improvement as underlying captioning models evolve.
\end{itemize}

\section{Related Work}
\noindent\textbf{Existing AVLLMs and OLMs.} AVLLMs are designed to accept audio and video input alongside language. Prominent open-source examples include ImageBind-LLM \cite{r:26}, ChatBridge \cite{r:25}, Macaw-LLM \cite{r:24}, VAST \cite{r:23}, OneLLM \cite{r:22}, VideoLLaMA2 \cite{r:10}, and video-SALMONN \cite{r:27}. More recently, OLMs have emerged that are capable of processing a mix of text, image, audio, and video input through a single interface. Key models in this space include Reka \cite{r:17}, Unified IO2 \cite{r:06}, MiniCPM-O2.6 \cite{r:09}, Ola \cite{r:05}, Baichuan Omni 1.5 \cite{r:08}, and Qwen2.5-Omni \cite{r:07}. While these models demonstrate impressive multimodal processing capabilities, their ability for deep, cross-modal reasoning remains a significant area for improvement.

\noindent\textbf{Reasoning capabilities in MLLMs.} The exploration of reasoning in language models was significantly advanced by Chain-of-Thought (CoT) prompting, which demonstrated that eliciting step-by-step reasoning improves performance on complex tasks~\cite{r:40}. This foundational work inspired more advanced techniques such as Self-Consistency, which samples multiple reasoning paths and selects the most consistent answer~\cite{r:41}, and Tree of Thoughts, which explores reasoning as a tree-search problem~\cite{r:42}. However, a critical challenge is ensuring the faithfulness of these generated reasoning chains, as studies have shown that models can produce plausible-looking explanations that do not reflect their actual inference process~\cite{r:43}. This concept of eliciting reasoning has been extended to the multi-modal domain with frameworks such as Multi-Modal Chain-of-Thought (MM-CoT), which generates reasoning chains that incorporate both textual and visual information~\cite{r:44}. Despite these advances in generating reasoning traces, the systematic evaluation of their factual grounding and logical validity, particularly for complex audio-visual tasks, remains an open problem that our work directly addresses.

\noindent\textbf{Existing Benchmarks for Evaluating AVLLMs and OLLMs.} Several benchmarks are used to evaluate the audio and visual processing capabilities of AVLLMs. Some of the popular AVLLM benchmarks are VGGSound~\cite{vggsound}, Music-AVQA~\cite{r:47}, AVQA~\cite{avqa}, and AVInstruct \cite{avinstruct}. Recently, several benchmarks have been proposed to evaluate OLMs' performance on a wider range of modalities, such as text, image, audio, and video. These benchmarks include Daily-Omni~\cite{r:03}, VideoVista~\cite{r:01}, CG-Bench~\cite{r:04}, and OmniEval~\cite{r:02}. Although these benchmarks evaluate the audio-visual capabilities of these models, they fail in evaluating deeper cognitive tasks such as causal reasoning, audio-visual synchronization, or the ability to identify implicit contradictions across modalities.

\section{\benchmarkname{}} \label{datasetPrep}
\subsection{Task Categories and Benchmark Statistics}
The \benchmarkname{} benchmark includes over 1,600 question–answer pairs across six cognitively distinct task categories: 
\textit{Cross-Modal Causal Reasoning (CR)} assesses a model's ability to infer cause-and-effect relationships between audio and visual events, \textit{Unanswerability (UANS)} evaluates a model's ability to identify and refuse to answer questions based on a false premise using both audio-visual cues, \textit{Timbre/Pitch Reasoning (TPR)} evaluates fine-grained auditory perception by requiring models to distinguish between the acoustic properties (e.g., timbre, pitch) of multiple sound sources and then link the target sound to a specific visual attribute, \textit{Tempo/AV Synchronization Analysis (TSA)} tests temporal alignment understanding, requires models to determine if audio and video streams are synchronized, \textit{Performer Skill Profiling (PSP)} evaluates a model's ability to make nuanced qualitative judgments, and \textit{Implicit Distractions (ID)} probes spatial attention and grounding. Each is designed to test a specific aspect of audio–visual understanding. Figure~\ref{fig:six_pngs} illustrates the benchmark and provides a paired question-answering example for each task. 

The number of multiple-choice questions in each task category, the percentage distribution of video samples by duration, and the contribution of samples based on their source of origin are shown in Fig~\ref{fig:distribution}(a), (b), and (c), respectively. Further implementation details are provided in the Appendix.

\begin{figure}[t]
    \centering
    \includegraphics[width=0.45\textwidth]{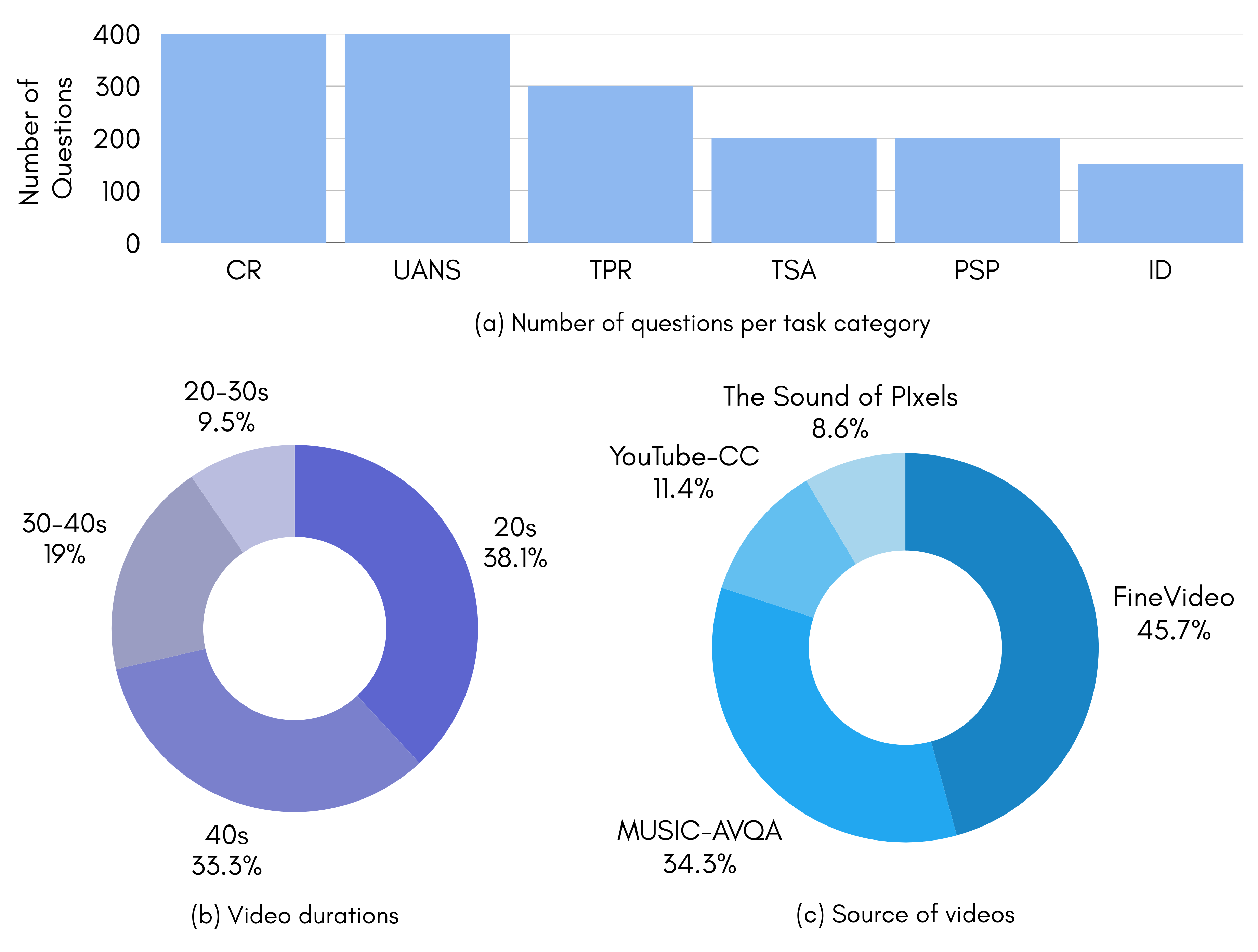}
    \caption{\textbf{Dataset statistics of \benchmarkname{}.} The plots in the above figure showcase the details of the data distribution in \benchmarkname{}. (a) The number of MCQ questions in each QA/task category. 
    (b) Percentage distribution of video samples in the dataset based on their duration. (c) Contribution of video samples to the dataset based on their source of origin.} 
    \label{fig:distribution}
\end{figure}



\begin{figure*}[t]
    \centering
    \includegraphics[width=\textwidth]{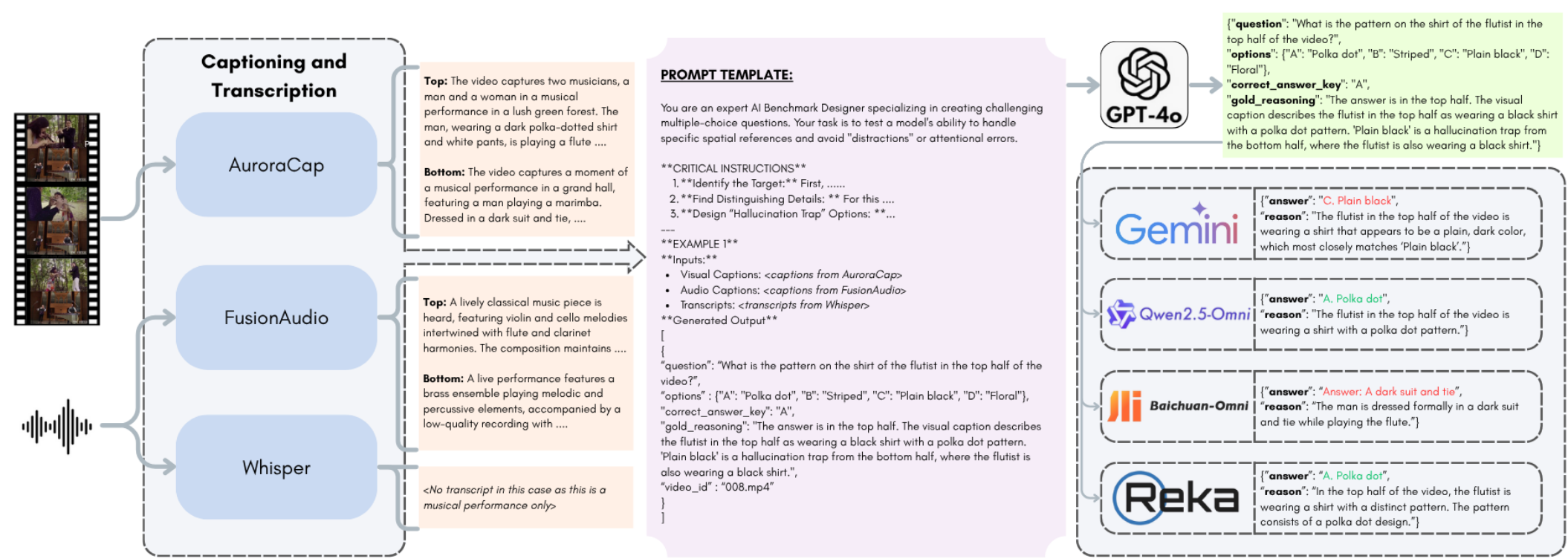}
    \caption{\textbf{The automated QA generation pipeline for the \benchmarkname{} benchmark.} The process begins with (1) Captioning and Transcription, where an input video is decomposed into text descriptions using specialized models. These multi-modal annotations are then inserted into a (2) Prompt Template for GPT-4o. The complete prompt is then used to (3) Generate a structured \texttt{json} output containing the question, multiple-choice options, the correct answer, and a "gold standard" reasoning trace.}
    \label{fig:qa_generation}
\end{figure*}

\subsection{QA Generation Pipeline}\label{QA pipeline Sect}
We developed a fully automated, multi-stage pipeline, illustrated in Figure \ref{fig:qa_generation}, to generate the QA pairs for \benchmarkname{}, ensuring both quality and scalability, consisting of three main stages:

    \noindent \textbf{Captioning and Transcription:} As shown on the left of Fig \ref{fig:qa_generation}, the process begins by generating rich descriptive text annotations for each modality from the input video. We employ AuroraCap for dense visual captions, FusionAudio for audio event captions, and Whisper to obtain speech transcripts. These multi-modal descriptions form the factual basis for the subsequent generation step.
    
    \noindent \textbf{Prompting and Generation:} The core of our pipeline involves feeding these annotations into a detailed \textit{Prompt Template} for our generator LLM, GPT-4o. For each of our six task categories, we developed a unique system prompt that outlines the specific rules and objectives, such as creating "hallucination traps" for the ID task or introducing a factual "perturbation" for the UANS task. As seen in Figure \ref{fig:qa_generation}, the prompt includes in-context examples to guide the model in generating a high-quality question, multiple-choice options, and the precise "gold reasoning" for the correct answer.
    
    \noindent \textbf{Structured Output and Validation:} Finally, the LLM is instructed to return a structured JSON object containing the complete QA pair, as depicted on the right of Fig \ref{fig:qa_generation}. Subsequently, our scripts parse this output, perform final validation checks, and append each question as a new line in a \texttt{jsonl} file, creating a clean, readable evaluation dataset.

\section{Evaluation and Metrics}
To rigorously assess the reasoning capabilities of OLMs on the \benchmarkname{} dataset, we move beyond simple answer accuracy. In multiple-choice question answering, a model can produce the correct answer using flawed or "unfaithful" reasoning ~\cite{r:30}. Therefore, our primary evaluation focuses on the quality of the generated reasoning trace. Inspired by recent studies on reasoning evaluation ~\cite{r:31}, we introduce \evalmetricname{} as a two-part decomposed scoring metric that separately measures the factual basis, \textit{Factual Consistency Score (FCS)}, and the logical validity of a model's explanation \textit{Core Inference Score (CIS)}, in addition to the \textit{Answer Accuracy (ACC)} (0 or 1). We present the evaluation pipeline in the Algorithm~\ref{alg:evaluation}.

\noindent\textbf{Factual Consistency Score (FCS):} The first component of our metric, FCS, evaluates whether the generated reasoning is factually grounded in the audio-visual context. A model's reasoning may be logically coherent but fail because it misidentifies key entities, attributes, or their locations. For example, in the TPR task, it might correctly infer that a violin has a higher pitch than a cello but mistakenly describes the violinist as a \textit{"man on the left"} instead of \textit{"woman on the right"}.

We employ a Large Language Model (LLM) as an evaluator to compute the FCS. For each response, we provide GPT-4o with both model-generated reasoning and ground-truth (GT) reasoning. The LLM is prompted with a detailed set of instructions to perform the following steps:

\begin{enumerate}
    \item \textit{Deconstruct Facts:} It first breaks down both reasoning strings into their core factual elements (e.g., subject, object, attribute, location).
    \item \textit{Compare Entities:} It then systematically compares the elements extracted from the generated reasoning with those from the GT reasoning.
    \item \textit{Compute Score:} The FCS is calculated as the fraction of factual elements of the GT reasoning that are correctly and completely represented in the generated reasoning. The score is a floating number between 0.0 and 1.0.
\end{enumerate}


\noindent\textbf{Core Inference Score (CIS):} The second component, CIS, is designed to evaluate the logical soundness of the reasoning, independent of its factual precision. A model could be factually correct about every detail, but still make a logically invalid inference. This score isolates and measures the quality of that abstract logical step. It is computed via a two-stage process that involves using LLM to "sanitize" the reasoning traces, e.g., to clean and simplify them, followed by a Natural Language Inference (NLI) check.

\begin{enumerate}
    \item \textit{Reasoning Sanitization:} We first employ GPT-4o to sanitize both the generated and GT reasoning traces. This process removes all specific, grounded details (e.g., "the man," "polka-dotted shirt," "in the top half") and distills each explanation into its most abstract logical claim (e.g., "the subject's attire has a specific pattern"). This ensures we are comparing the core logic, not the factual details already assessed by the FCS.
    \item \textit{NLI-based Scoring:} The sanitized GT reasoning (as the premise) and the sanitized generated reasoning (as the hypothesis) are input to a pre-trained cross-encoder NLI model (\texttt{nli-deberta-v3-base}) from Hugging Face. We choose an NLI model, rather than a semantic similarity model, because it is specifically trained to assess logical entailment. It determines if the premise logically implies the hypothesis, which is a stricter and more appropriate measure for evaluating reasoning. The final CIS is the entailment probability score produced by the model, a floating number between 0.0 and 1.0.
\end{enumerate}
By proposing these two scores, our \evalmetricname{} provides a robust evaluation of a model's ability to not only perceive audio-visual content accurately but also to reason coherently about what it perceives.


\tikzstyle{block} = [rectangle, draw, text centered, minimum height=1.5em, font=\footnotesize, rounded corners]
\tikzstyle{arrow} = [->, thick]

\begin{algorithm}[t]
\caption{Evaluation Procedure for \evalmetricname{}}
\label{alg:evaluation}
\begin{algorithmic}[1]
\State \textbf{Input:} Question $Q$, Model Answer $A$, Model Reasoning $R_{\text{model}}$, Ground-Truth Reasoning $R_{\text{gt}}$
\State \textbf{Output:} Evaluation Score $\in [0, 1]$
\If{$A$ is not correct}
    \State \Return 0
\Else
    \State \textbf{/* Split Evaluation */}
    \State \textbf{Factual Consistency Score (FCS):}
    \State \hspace{0.5cm} 1. Deconstruct $R_{\text{model}}$ and $R_{\text{gt}}$ into atomic facts
    \State \hspace{0.5cm} 2. Compare factual entities and relations
    \State \hspace{0.5cm} 3. Compute FCS $\in [0, 1]$ as fraction of matching facts
    \State \textbf{Core Inference Score (CIS):}
    \State \hspace{0.5cm} 1. Sanitize $R_{\text{model}}$ to isolate reasoning content
    \State \hspace{0.5cm} 2. Use NLI-based scoring to compare with $R_{\text{gt}}$
    \State \hspace{0.5cm} 3. Compute CIS $\in [0, 1]$ as entailment probability
    \State \Return FCS, CIS 
\EndIf
\end{algorithmic}
\end{algorithm}

\section{Experiments, Results and Analysis}
\subsection{Model Selection and Evaluation Scores}
We selected the SOTA models OneLLM~\cite{r:22}, VideoLLaMA~\cite{r:49}, VideoLLaMA2~\cite{r:10}, \textit{Ola}~\cite{r:05}, \textit{UnifiedIO2}~\cite{r:06}, \textit{Qwen 2.5 Omni}~\cite{r:07}, and \textit{Baichuan Omni 1.5}~\cite{r:08} as open-source models, and \textit{Reka}~\cite{r:17}, \textit{Gemini 2 Flash}, and \textit{Gemini 2.5 Flash} as closed-source models to evaluate on our \benchmarkname{} dataset. For each model, we post-processed the raw reasoning output to extract its selected answer for every question (see Section \ref{datasetPrep} for details on our QA formatting across proposed tasks).

\subsection{Performance Analysis of the Models}

We present the evaluation results of the SOTA AVLLMs and OLMs using key metrics across the $6$ tasks on \benchmarkname{} dataset in the Table \ref{omnimodalsCmp}, which summarizes the performance of evaluated models across three metrics: ACC, FCS, and CIS. We report the key findings of the models' performance for each task as follows:

\begin{table*}[t]
\centering
\small 
\renewcommand{\arraystretch}{1.0}
\resizebox{\textwidth}{!}{
\begin{tabular}{@{}l|ccc|ccc|ccc|ccc|ccc|ccc@{}}
\toprule
& \multicolumn{3}{c|}{\cellcolor[gray]{0.8}{\textbf{Distractions}}} & \multicolumn{3}{c}{\cellcolor[gray]{0.8}{\textbf{Causal Reasoning}}} & \multicolumn{3}{c}{\cellcolor[gray]{0.8}{\textbf{Skill Profiling}}} & \multicolumn{3}{c}{\cellcolor[gray]{0.8}{\textbf{Timbre/Pitch Reasoning}}} & \multicolumn{3}{c}{\cellcolor[gray]{0.8}{\textbf{Unanswerability}}} & \multicolumn{3}{c}{\cellcolor[gray]{0.8}{\textbf{Tempo/AV-Sync.}}} \\
\cmidrule(lr){2-4} \cmidrule(lr){5-7} \cmidrule(lr){8-10} \cmidrule(lr){11-13}\cmidrule(lr){14-16}\cmidrule(lr){17-19}
\textbf{Method} &  ACC & FCS & CIS & ACC & FCS & CIS &  ACC & FCS & CIS &  ACC & FCS & CIS &  ACC & FCS & CIS&  ACC & FCS & CIS\\
\midrule
Qwen2.5-Omni 7B \cite{r:07} & \bf 82.67 & \bf 67.95 & \bf 45.23 & \bf 80 & \bf 45.5 & \bf 32.3 & \bf73 & \bf 28.57 & \bf27.96 & \bf 72.09 & \bf 42.44 & \bf 29.98 &	\bf92 & \bf 68 & \bf 61.03 & \bf 44 & \bf 29.5 & 15.68 \\
Qwen2.5-Omni 3B \cite{r:07} & 80 & 53.78 & \bf 37.11 & \bf73.03 & 41.57 & \bf 31.57 & 69.23 & 24.78 & 20.51 & \bf63.73 & \bf38.05 & \bf 19.34 & \bf88.33 & \bf63.33 & \bf 60.2 & 37.5 & 21.88 & 14.62 \\
Unified-IO 2 \cite{r:06} & 23.23 & 8.57 & 15.05 &	44.44 &	21.91 &	22.74 &	57.93 &	11.60 &	\bf28.95 & 28.36 & 7.21 & 13.86 &	72.73 &	55.45 & 50.60 & 24.14 & 13.79 & \bf 20.71 \\
Baichuan-Omni 1.5 \cite{r:08} &	10.00 & 0.00 & 0.00 &	6 &	1.5 & 3.48 & 20 & 3.75 & 0.44 &	18.75 &	4.17 & 2.07 & 8.16 & 0.00 & 0.00 & 30 &	2.5 & 3.97 \\
Ola \cite{r:05}	& 63.33 & 52.71 & 24.21 &	67.28 &	44.39 & 21.36 &	34.50 &	15.59 &	19.54 &	55.39 &	29.53 &	\bf 19.68 &	\bf 85.42 &	\bf 73.46 &	\bf 61.10 &	37.11 &	26.16 &	15.31\\
\cmidrule{1-19}
OneLLM \cite{r:22}	& 24.59 & 7.38 & 5.73 & 26.67 & 12.92 & 6.82 & 28.57 & 7.86 & 11.48 & 21.57 & 5.39 & 5.03 & 15.52 & 8.19 & 1.82 & 34.43 & 21.72 & 14.81 \\
VideoLLaMA \cite{r:49}	& 42.07 & 13.60 & 11.32 & 58.00 & 27.24 & 20.11 & 51.71 & 20.13 & 10.69 & 55.43 & 18.82 & 15.65 & 25.27 & 10.41 & 10.67 & 30.35 & 15.23 & 12.88 \\
VideoLLaMA2 \cite{r:10}	& 46.00 & 22.72 & 10.90 & 65.00 & 30.50 & 25.20 & 53.00 & 18.00 & 14.70 & 56.00 &	21.50 &	18.14 &	22.00 & 11.45 & 13.94 & 38.00 &	19.50 & 18.81 \\
\cmidrule{1-19}
\demph{Reka \cite{r:17}}	& \demph{40.67}& 
\demph{30.33}&	
\demph{13.01}&	
\demph{66.16}&	
\demph{\bf 46.34}&	
\demph{\bf 24.11} &	
\demph{11.62}&	
\demph{1.52} &	
\demph{1.00} &	
\demph{33.84} &	
\demph{10.48} &	
\demph{5.90} &	
\demph{67.17} &	
\demph{47.65} &	
\demph{25.59} &	
\demph{41.41}&	
\demph{26.77} &	
\demph{10.50} \\
\demph{Gemini 2.0 Flash} & 
\demph{\bf 81.33} & 
\demph{\bf 64.12} & 
\demph{\bf36.33} &
\demph{\bf80} & 
\demph{44.5} & 
\demph{22.64} &
\demph{\bf74} &
\demph{\bf31.64} &
\demph{\bf33.38} &
\demph{\bf 72} &
\demph{\bf 47.5} & 
\demph{\bf 37.09} &
\demph{62} & 
\demph{31} &
\demph{22.03} &
\demph{\bf50} & 
\demph{\bf29} &	
\demph{\bf23.88} \\
\demph{Gemini 2.5 Flash} &	
\demph{\bf 84.55} &	
\demph{\bf65.23} & 
\demph{18.77} & 
\demph{\bf76.36} & 
\demph{\bf45.18} &	
\demph{21.11} &	
\demph{\bf85.45} &	
\demph{\bf52.00} &	
\demph{17.61} &	
\demph{62.73} &
\demph{36.32} &	
\demph{13.92} &	
\demph{52.73} &	
\demph{26.23} &	
\demph{23.27}& 
\demph{\bf44.55} &	
\demph{\bf30.95} &	
\demph{\bf25.78}\\

\bottomrule
\end{tabular}}

\caption{\textbf{Evaluation Scores for Omni-modal LLMs.} Our proposed \evalmetricname{}, consisting of ACC(Accuracy), FCS (Factual Consistency Score), and CIS (Core Inference Score), provides insight into the performance of the SOTA Models across multiple tasks in \benchmarkname{}. \demph{Closed source models are at the bottom.}}
\label{omnimodalsCmp}
\end{table*}

\noindent \textbf{Timbre and Pitch Reasoning.} Among the evaluated models, Gemini 2.0 Flash and Qwen2.5-Omni-7B demonstrate strong overall performance, with ACC scores of 72.0\% and 72.09\%, respectively. While these high scores reflect a solid capability in timbre and pitch classification, a deeper look at the supporting metrics, FCS and CIS, reveals noticeable differences in reasoning quality. In particular, Gemini 2.0 Flash outperforms all others in both FCS (47.5\%) and CIS (37.09\%), suggesting it not only classifies correctly but also grounds its reasoning more effectively in perceptual evidence and follows more logically coherent inference steps. In contrast, while Qwen2.5-Omni-7B matches Gemini in accuracy, its lower FCS (42.44\%) and CIS (29.98\%) suggest that correct predictions may sometimes result from shortcut learning or shallow heuristics rather than sound reasoning. The smaller variant Qwen2.5-Omni-3B performs moderately across all metrics, with reduced capacity likely contributing to lower consistency and inference quality. 

These results highlight a broader challenge: reasoning about timbre and pitch requires models not only to recognize auditory features but to map them onto a chain of logically valid, perceptually grounded steps. This task appears to be particularly demanding for smaller models, likely due to limitations in their representation depth and reasoning robustness. The superior performance of Gemini 2.0 Flash across all metrics points to architectural or training differences that better support reasoning over auditory features. This may include enhanced perceptual grounding, better alignment between AV representations, or more effective use of instruction tuning.


\noindent \textbf{Tempo and Audio-Visual Synchronization.} This task proves especially challenging for all models, with relatively low accuracy across the board, indicating inherent difficulty in aligning temporal auditory cues with visual information. The best-performing models, Gemini 2.0 Flash (55\% ACC) and Gemini 2.5 Flash (44.5\% ACC), suggest that closed-source Gemini models are better equipped for this kind of multi-modal temporal reasoning, potentially owing to more sophisticated alignment mechanisms or larger, more specialized training corpora. Interestingly, while Gemini 2.0 Flash leads in overall ACC, Gemini 2.5 Flash achieves the highest FCS (30.95\%) and CIS (25.78\%), indicating that it produces more perceptually grounded and logically coherent reasoning steps despite lower raw accuracy. The relatively close performance of Qwen2.5-Omni-7B (44\% ACC) suggests that open models can approach competitive levels in raw prediction, but their slightly weaker FCS and CIS highlight a gap in reasoning fidelity. 

These findings underscore the complexity of tempo and synchronization tasks, which likely demand precise temporal modeling and cross-modal understanding capabilities that are not yet fully developed in smaller or open-source models. Future research could explore ways to improve temporal alignment learning, possibly through pretraining with synchronized AV data or designing inductive biases that promote rhythm-sensitive representations. Ultimately, while Gemini models currently set the performance upper bound, low absolute scores across all metrics suggest substantial room for progress in this domain.

\noindent\textbf{Performer Skill Profiling.} This task challenges models to make fine-grained multi-modal judgments, requiring them to integrate subtle visual and auditory cues to distinguish novice from expert performers. In this setting, Gemini 2.5 Flash achieves the highest overall accuracy (84.45\%), followed by Gemini 2.0 Flash (74\%), and open-source models like Qwen2.5-Omni-7B (73\%) and Qwen2.5-Omni-3B (69.23\%). However, a closer look at the reasoning metrics reveals key differences in how these models arrive at their predictions. Gemini 2.5 Flash excels in factual consistency (FCS: 52.00\%), indicating strong grounding in perceptual evidence—likely due to better multi-modal representation learning—while Gemini 2.0 Flash leads in core inference score (CIS: 33.39\%), suggesting more coherent reasoning chains. Open-source models such as Qwen2.5 and Unified-IO2 lag in both FCS and CIS, which may reflect weaker integration across modalities or limited exposure to expert-novice skill contrasts during training. 

These findings highlight that PSP is not merely a classification task, but one that stresses deep reasoning grounded in perceptual subtleties - an area where current models, especially open-source ones, still struggle. The superior performance of Gemini models may be attributed to larger or better-aligned pretraining. Future work should explore targeted fine-tuning on expert-novice contrastive data and improved reasoning supervision to close this performance gap and better equip open models for nuanced AV understanding.


\noindent\textbf{Cross-Modal Causal Reasoning.} This task challenges models to detect cause-and-effect relationships that span auditory and visual modalities, such as linking a specific sound (e.g., a blender whirring) to its visual consequence (e.g., fruit being pureed). Success requires more than unimodal perception; it demands temporal and semantic alignment across modalities. In this setting, Qwen2.5-Omni-7B achieves the highest accuracy (80\%), indicating strong raw prediction ability, closely followed by Gemini 2.5 Flash (76.36\%) and Qwen2.5-Omni-3B (73.03\%). Interestingly, Reka, despite a lower overall accuracy, demonstrates the strongest factual consistency (FCS: 46.34\%), suggesting that its reasoning is more consistently grounded in actual causal mechanisms—even if the final answer is sometimes incorrect. Qwen2.5-Omni-7B leads in core inference score (CIS: 32.3\%), pointing to more coherent step-by-step reasoning, with Qwen2.5-Omni-3B (31.57\%) and Reka (24.11\%) trailing.

These results point to a trade-off between accuracy and depth of reasoning: while Qwen2.5-Omni models are better at getting the \textit{right answer}, Reka may be better at understanding \textit{why} the answer makes sense, especially when aligning causality across modalities. This suggests current models still struggle to unify multi-modal streams in a causally robust way. Future improvements could focus on refining temporal modeling and explicitly training models on cross-modal causal structures, rather than relying purely on correlation-based associations.

\noindent\textbf{Implicit Distractions.}  
Gemini 2.5 Flash achieves the highest accuracy (84.55\%), closely followed by Qwen2.5-Omni-7B (82.67\%) and Gemini 2.0 Flash (81.33\%), suggesting that all three models are generally capable of resolving ambiguous referents. However, Qwen2.5-Omni-7B stands out in its ability to justify its decisions: it leads in factual consistency (FCS: 67.95\%) and core inference score (CIS: 45.23\%), outperforming Gemini 2.5 Flash (FCS: 65.23\%, CIS: 37.11\%) and Gemini 2.0 Flash (FCS: 64.12\%, CIS: 36.33\%). 

This indicates that Qwen2.5-Omni 7B not only gets the answer right but does so through more coherent and perceptually grounded reasoning. Its success may stem from stronger spatial grounding capabilities or a pretraining regime that emphasizes attention to referential cues in multi-modal contexts. On the contrary, the Gemini models, while highly accurate, occasionally exhibit less transparent reasoning, which may reflect over-reliance on shortcut heuristics. These findings underscore the subtle complexity of distinguishing between visually similar entities under linguistic constraints, an ability not uniformly mastered across models. Future work could benefit from incorporating explicit spatial grounding objectives or contrastive learning setups that train models to resist distractor interference.


\noindent\textbf{Unanswerability.} 
This task requires an ability closely tied to models' honesty and epistemic uncertainty, demanding not just perceptual grounding but also cognitive awareness: recognizing when the multi-modal content contradicts the question. In this setting, Qwen2.5-Omni-7B achieves the highest accuracy (92\%), with Qwen2.5-Omni-3B (88.33\%) and Ola (85.42\%) following closely. These high scores suggest that models are generally capable of identifying contradictions in video content. However, when evaluating the quality of reasoning behind those decisions, Ola stands out. It achieves the highest factual consistency (FCS: 73.46\%) and core inference score (CIS: 61.10\%), indicating that its rejections are not just correct, but also supported by well-grounded and logically sound justifications. Qwen2.5-Omni-7B and 3B are close in CIS (61.03\% and 60.2\%) but show slightly lower FCS (68\% and 63.33\%), suggesting more occasional reliance on heuristic cues or pattern matching rather than fully grounded reasoning. 

Ola’s strength may lie in better calibration, demonstrating confidence when appropriate, and hesitation when needed, which is essential for trustworthy AI in open-ended settings. These findings highlight the subtlety of this task: high accuracy alone does not imply genuine understanding unless paired with consistent, explainable reasoning. Future work might explore training objectives that reward epistemic humility, as well as incorporating counterfactual or adversarial data that explicitly probe a model’s ability to detect and reject misinformation.



\section{Limitations and Future Work}
While \benchmarkname{} and \evalmetricname{} introduce a new paradigm for fine-grained reasoning evaluation, they have limitations that offer clear paths for future research. First, the benchmark, though centered on six challenging cognitive tasks, could be broadened to include more reasoning types and diverse video content. Evaluating models on long-form narratives or skills in areas like sports or surgery would help assess generalizability. Second, \evalmetricname{} depends on a proprietary LLM (GPT-4o) to assess Factual Consistency (FCS), posing challenges due to the model's stochasticity, potential biases, and limited accessibility, hindering reproducibility and broader community adoption. Future directions include: \textit{(i) Scaling and Diversifying the \benchmarkname{} Benchmark:} our automated, modular QA pipeline supports scaling AURA with additional data and task categories, including videos from varied sources to improve model robustness and generalization. \textit{(ii) Developing Robust and Open-Source Evaluators:} to reduce reliance on proprietary LLMs and enhance reproducibility, focusing on building open-source models tailored to evaluate reasoning trace faithfulness. This would lower costs, improve accessibility, and support more transparent and controlled evaluation of factual grounding and logical coherence.

\section{Conclusion}
We proposed \benchmarkname{}, the first benchmark that enables us to evaluate the cross-modal, audio-visual, reasoning performance of the SOTA models on the $6$ fine-grained cognitive tasks. We used the automated approach to generate this benchmark, which ensures the scalability of the dataset as well. We present \evalmetricname, which consists of FCS and CIS as the key components for evaluating the reasoning ability of the models. This helped us to understand the meaningful differences among the models' factual alignment and inference abilities. To summarize, our benchmark provides a valuable foundation for a holistic AV evaluation, fostering deeper insights into model behavior across AV reasoning tasks.




\twocolumn[
\begin{center}
{\LARGE\bfseries
AURA: A Fine-Grained Benchmark and Decomposed Metric for Audio-Visual Reasoning\par}
\vspace{0.5em}
{\Large\textcolor{blue}{Supplementary Material}\par}
\vspace{1em}
\end{center}
]

\noindent \textbf{Supplementary Material organized as follows:}
\begin{enumerate}[label=\Alph*]
  \item \textbf{Additional Details on AURA}
  \item \textbf{GPT/Model Prompts}
  \item \textbf{Implementation Details}
  \item \textbf{More Qualitative Examples}
  \item \textbf{Additional Related Works}
\end{enumerate}

\appendix 
\section{Additional Details on AURA}
\subsection{Design Principles}
The design of \benchmarkname{} is guided by three core principles to overcome these gaps:

\begin{enumerate}
\item \textbf{Mandatory Cross-Modal Reasoning:} Every question is constructed such that it is unanswerable using only a single modality (audio or visual). Models are forced to locate, integrate, and reason about complementary clues from both the audio and visual streams, preventing reliance on uni-modal heuristics.
\item \textbf{Advanced Cognitive Tasks:} We move beyond standard QA tasks to introduce novel categories that probe more nuanced cognitive abilities. These include discerning subtle acoustic properties like timbre and pitch, understanding audio-visual causality, detecting desynchronization, and identifying factual inconsistencies, which are often overlooked in other benchmarks.
\item \textbf{Automated and Scalable Generation:} We employ a modular, prompt-driven pipeline that leverages a state-of-the-art LLM (GPT-4o in our case) to generate high-quality, task-specific QA pairs. This approach ensures not only consistency and scalability but also allows for the creation of sophisticated linguistic structures and distractor options for each question.
\end{enumerate}

\subsection{Construction of Task Categories} \label{sec:task}
\noindent \textbf{Cross-Modal Causal Reasoning (CR):} This task assesses a model's ability to infer cause-and-effect relationships between audio and visual events. Questions require the model to identify the audio event that caused a visual outcome (e.g., the sound of a blender corresponding to fruit being pureed) or vice-versa. We source videos for this task from the FineVideo dataset~\cite{r:46}, as its focus on real-world scenarios provides a rich diversity of natural causal events.

\noindent \textbf{Unanswerability (UANS):} The formulation of this task was inspired by recent work on model honesty~\cite{r:45}. While the authors in this case had made assessments in the visual domain of videos, the UANS task in \benchmarkname{} evaluates a model's ability to identify and refuse to answer questions based on a false premise, using both audio-visual cues. Due to the cross-modal dependency of the task, the false premises and distracting choices are present in different modalities, making the questions much tougher. Questions contain a deliberate factual error regarding the video's content, and the model must recognize this contradiction to explain why the question is unanswerable. For this, we also leverage the diversity of the FineVideo dataset, which allows for the creation of plausible yet factually incorrect premises across a wide range of contexts.

\noindent \textbf{Timbre/Pitch Reasoning (TPR):} This task evaluates fine-grained auditory perception by requiring models to distinguish between the acoustic properties (e.g., timbre, pitch) of multiple sound sources and then link the target sound to a specific visual attribute. We source videos from the MUSIC-AVQA dataset~\cite{r:47}, which is ideal due to its wide variety of musical instruments, offering the diverse timbral properties and pitch ranges essential for this analysis.

\noindent \textbf{Tempo/AV Synchronization Analysis (TSA):} To test temporal alignment understanding, this task requires models to determine if audio and video streams are synchronized. By presenting a balanced set of both aligned and deliberately misaligned clips, we evaluate the ability to detect subtle desynchronization between a visual action and its sound. We use the MUSIC-AVQA dataset, as the structured nature of musical performances provides a clear and reliable basis for creating and evaluating these temporal misalignments.

\noindent \textbf{Performer Skill Profiling (PSP):} This task evaluates a model's ability to make nuanced qualitative judgments. By presenting sequential clips of novice and expert performers, we challenge the model to synthesize subtle visual cues (e.g., posture, finger fluidity) with auditory qualities (e.g., rhythmic stability, tonal clarity) to correctly identify each performer's skill level. Given the novelty of this task, we curated a custom set of videos from YouTube, sourced under Creative Commons (CC) licenses or with explicit creator permission, to ensure clear examples of both novice and expert performances.

\noindent \textbf{Implicit Distractions (ID):} This task probes spatial attention and grounding using split-screen videos. Models must answer questions about a "target" object in one specified region (e.g., "the guitarist at the top") while ignoring a similar distractor object in the other half. We utilize the MUSIC dataset~\cite{r:48} for this task, as its collection of duet performances provides a natural source for creating these controlled scenarios with plausible distractors.

\subsection{Benchmark Statistics}
The \benchmarkname{} benchmark comprises more than 1,600 pairs of question-answering, distributed in our six advanced cognitive task categories. The distribution is well-balanced, with Cross-Modal Causal Reasoning (CR) and Unanswerability (UANS) forming the largest categories with approximately 400 questions each, ensuring a robust evaluation of these fundamental reasoning skills.

Our data is curated from a diverse set of sources, with each source chosen to meet the specific requirements of our tasks. The majority of our videos (45.7\%) are sourced from the FineVideo dataset for the CR and UANS tasks. To focus on specific, isolated events, we process these by extracting 20-second clips based on activity levels and scene dynamism scores available in the metadata. For tasks requiring longer temporal context and fine-grained audio analysis, such as Timbre/Pitch Reasoning (TPR) and Tempo/AV Synchronization (TSA), we utilize the MUSIC-AVQA dataset (34.3\%), from which we trim 40-second clips.

To construct the Implicit Distractions (ID) task, we source duet performance videos from The Sound of Pixels dataset (8.6\%). For each sample, two clips containing a common instrument are stacked vertically to create a split-screen video, designed to probe the model's spatial awareness. Finally, for the novel Performer Skill Profiling (PSP) task, we curated a custom dataset (11.4\%) from YouTube videos with appropriate licenses and permissions. These 30-40 second clips, often from vlogs documenting a musician's learning journey, are concatenated to present a novice and an expert performance sequentially in a random order. This targeted data processing results in a varied distribution of video durations. The lengths are deliberately chosen to match task complexity, ranging from shorter 20-second clips (38.1\%) for causal events to longer 40-second clips (33.3\%) for tasks demanding sustained temporal analysis.

\section{GPT/Model prompts}
\begin{figure}[h!]
    \centering
    \includegraphics[width=\linewidth]{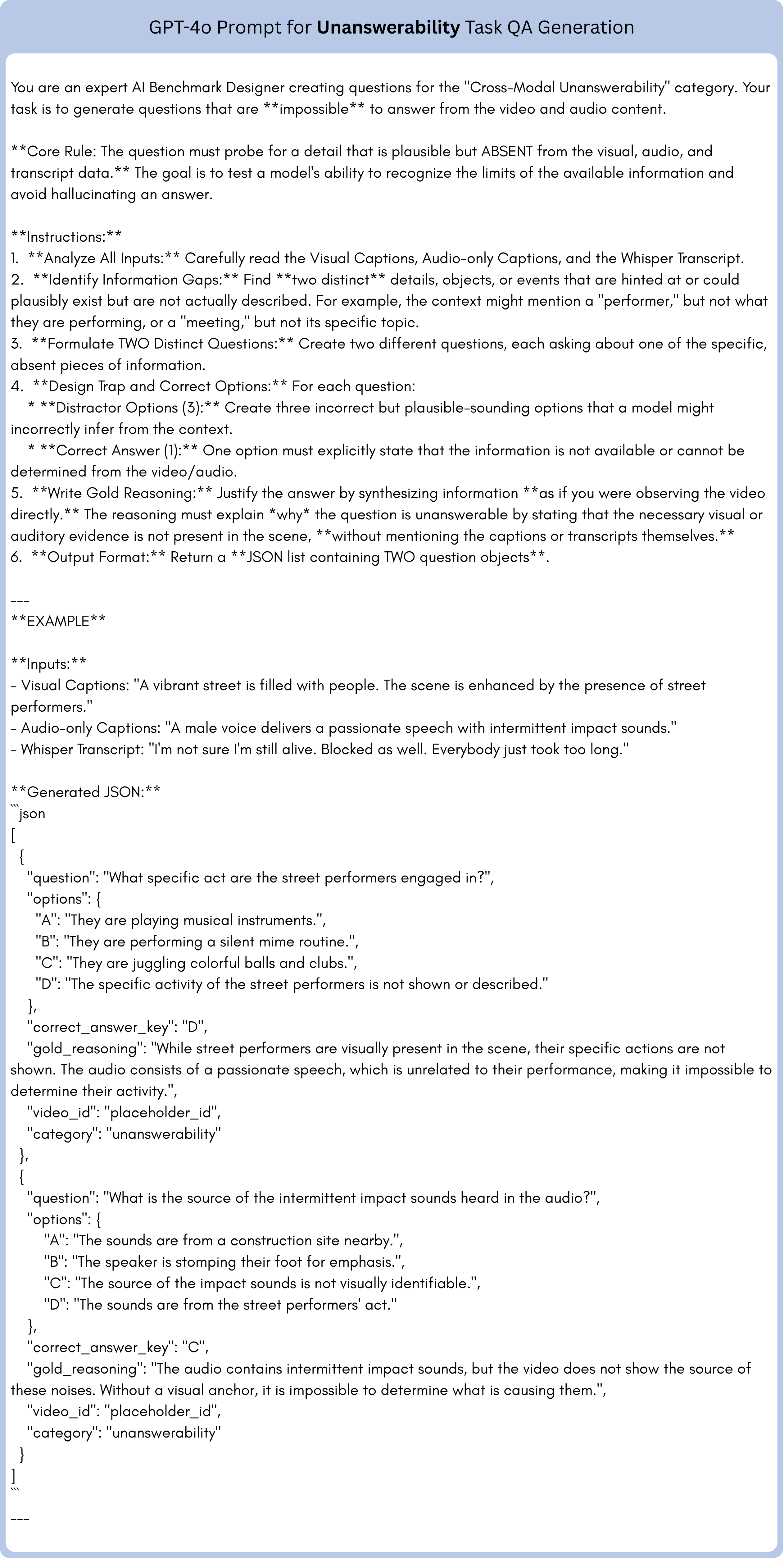}
    \caption{\textbf{QA Generation Prompt.} This prompt is used to guide the LLM in generating QA pairs of a certain task category. Each category has some Core Rules and Instructions along with in-context examples.}
    \label{fig:gpt_prompt_qa_gen}
\end{figure}

\begin{figure}[h!]
    \centering
    \includegraphics[width=\linewidth]{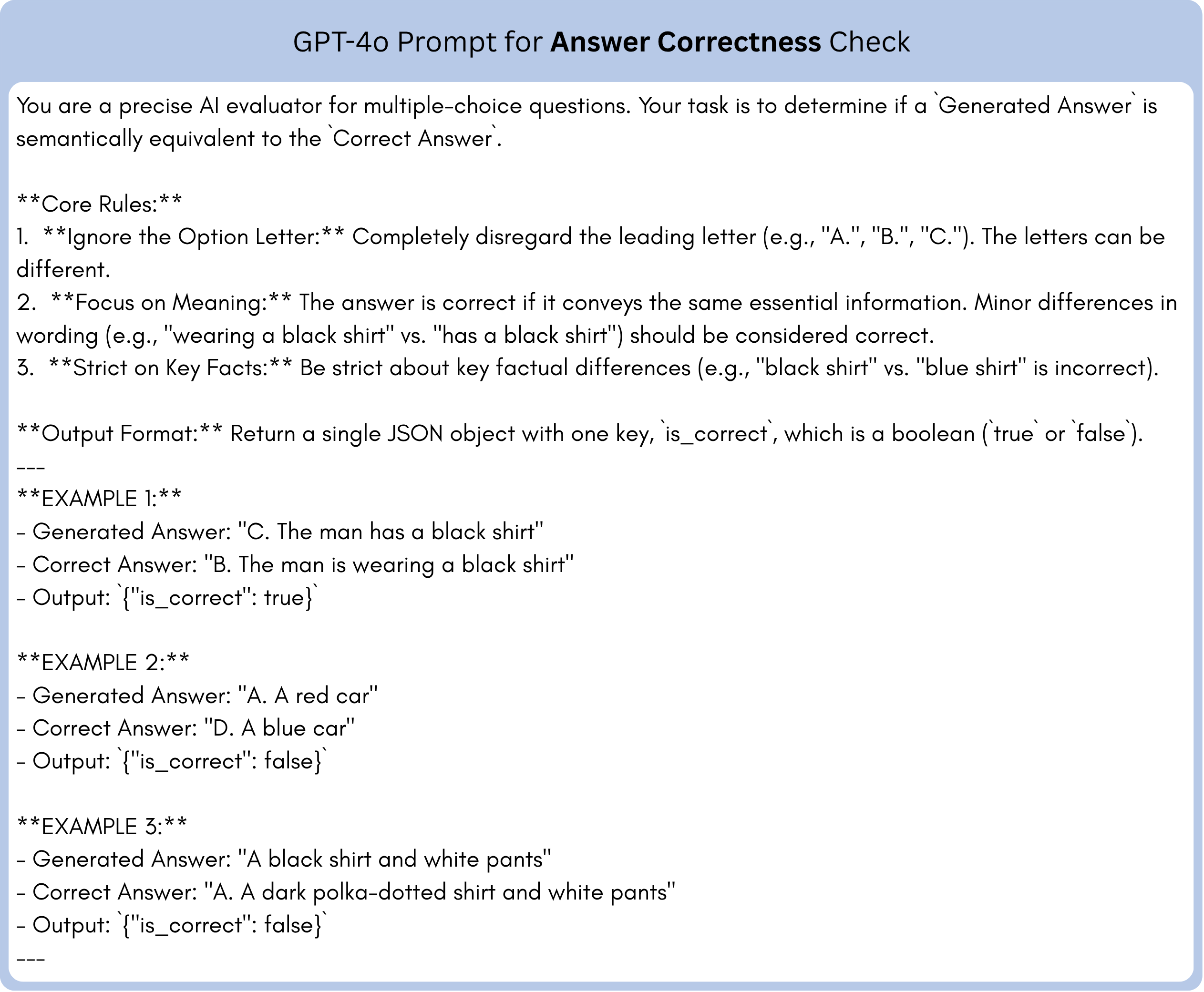}
    \caption{\textbf{Answer Correctness.} This prompt takes in both the GT MCQ-answer and the model-generated response and returns a binary value of correctness.}
    \label{fig:gpt_prompt_ans_crt}
\end{figure}

\begin{figure}[h!]
    \centering
    \includegraphics[width=\linewidth]{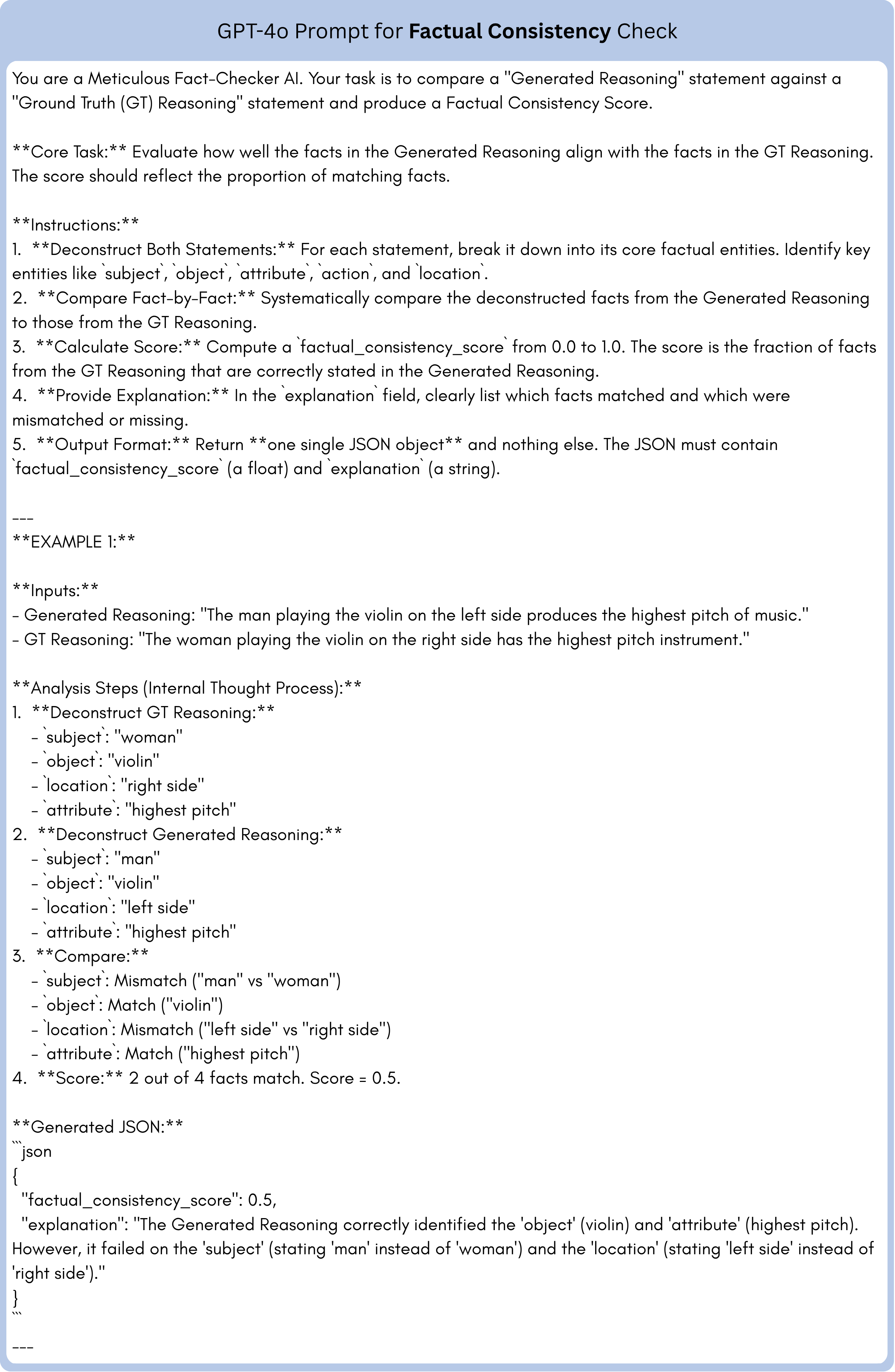}
    \caption{\textbf{Factual Consistency.} This prompt takes in both the GT and model-generated reasoning trace and matches only the factual objects. It then returns an accuracy score for the matching.}
    \label{fig:gpt_prompt_factual}
\end{figure}

We have used carefully structured prompts guiding GPT-4o in generating QA-pairs and calculating the \textit{Factual Consistency Score} (FCS) and \textit{Core Inference Score} (CIS). All the prompts contain a set of \textbf{Core Rules} that the LLM needs to strictly adhere to, along with in-context examples of the inputs and expected outputs. The first sample shown in Figure~\ref{fig:gpt_prompt_qa_gen} was used to generate QA-pairs for the Unanswerability task as explained in Section~\ref{sec:task}. The second sample shown in Figure~\ref{fig:gpt_prompt_ans_crt} was used to determine the \textit{answer correctness} of the model response when compared with the GT answer. And finally, we used the sample prompt shown in Figure~\ref{fig:gpt_prompt_factual} to calculate the factual consistency score of the reasoning traces.

\section{Implementation Details}
We use both open-source models and closed-source ones. All these models are processing the full videos natively, and we do not need to extract multiple frames for the video-audio question answering evaluations. To evaluate their performance, we generate each model's raw responses to each question and compare them with the ground-truth (human-level) reasoning generated by the proposed automated QA generation pipeline for the \benchmarkname{} benchmark. For the open-source model, we followed step by step the instructions that the authors provided in their GitHub repository, and for the closed-source models, we tried their official APIs, and we have evaluated the open-source models by 1 RTXA5000 GPU.

\section{More Qualitative Examples}
We present more qualitative samples for each task. As shown, Gemini 2.0 Flash, Gemini 2.5 Flash, and Qwen 7B are the top 3 models that have the best performance across the different tasks. While we had presented a summary of our results in the main paper, this is a more detailed analysis of some of the interesting observations that we made during the course of our evaluation and testing. The video data for these specific cases has been provided in the \texttt{zip} file, named as per the relevant task category.

We start with the observations we made during our evaluation of SOTA models on the \textit{Performer Skill Profiling} (PSP) and \textit{Implicit Distractions} (ID) task. For the PSP, we observe that while some of the models get the answer correct as to which of the two performers shown in the video is the \textit{expert}, all models except Gemini 2.5 Flash get the reasoning wrong. Gemini 2.5 Flash can not only identify the correct instrument being played by the expert, as has been asked in the question, but is also able to temporally localize the performance of the expert performer. This shows strong grounding and reasoning capabilities. All other models that get the answer correct state reasons which are more or less equivalent to \textit{"the expert performer is playing the guitar"}. This logic does not address \textbf{WHY} the model thought that the performer playing the guitar was the expert. Ideally, the model should have linked perceptible visual and audio attributes that correspond to a person's skill level and expertise, similar to what Gemini 2.5 Flash has done. Out of the 11 models that we tested, only 1 got both the reasoning correct. All the other models showed failures in some way or the other. The summary of the model performances of the PSP task has been shown in Table~\ref{tab:psp}.

\begin{table}[h!]
    \centering
    \begin{tabular}{lc}
    \toprule
    \textbf{Model Performance} & \textbf{Count} \\
    \midrule
    Answer + Reasoning Correct & 1 \\
    Answer Correct, Reasoning Wrong & 6 \\
    Answer Wrong & 4 \\
    \bottomrule
    \end{tabular}
    \caption{\textbf{PSP:} Model Performance Summary}
    \label{tab:psp}
\end{table}

\begin{figure*}[t!]
    \centering
    \includegraphics[width=\linewidth]{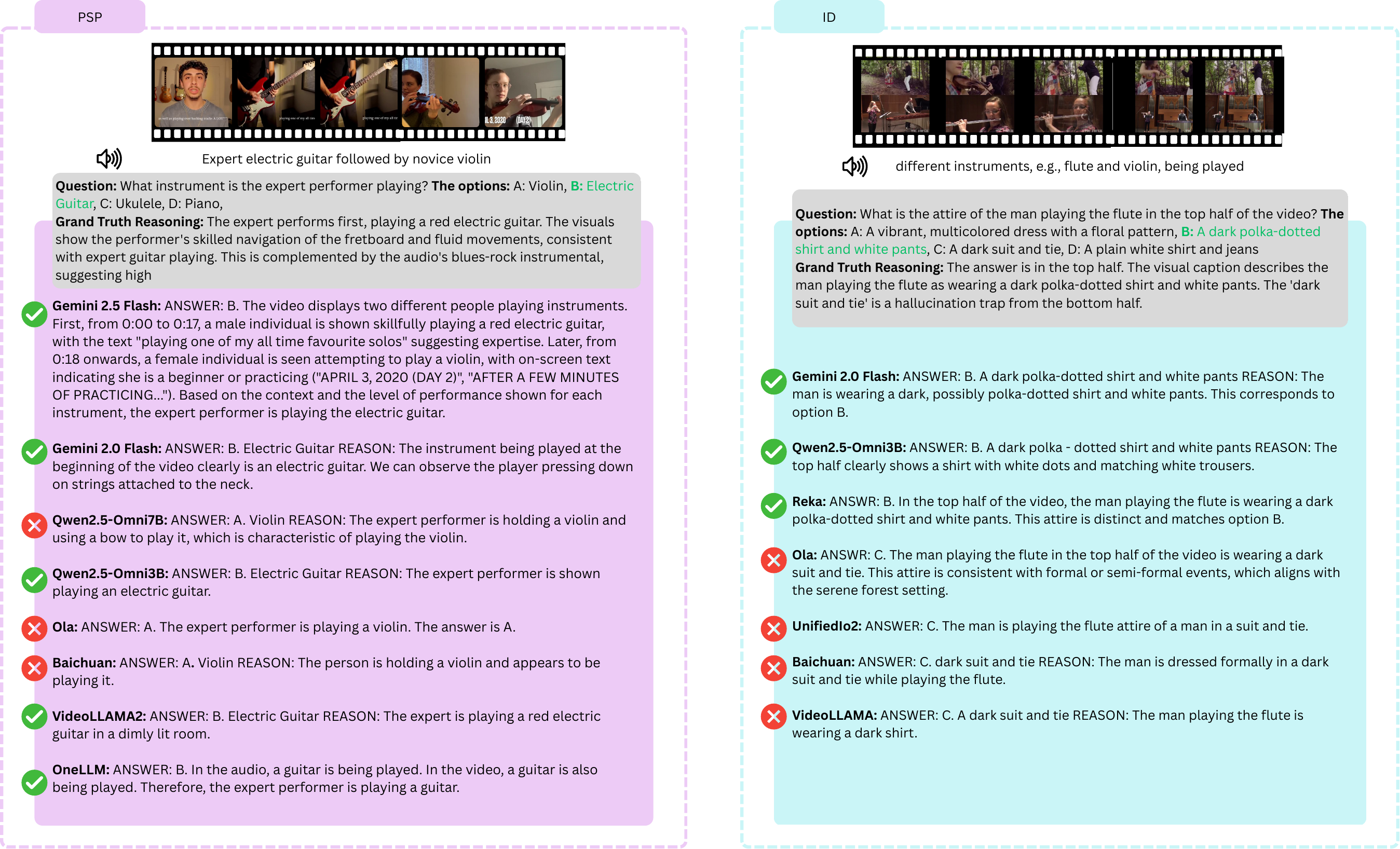}
    \caption{\textbf{PSP \& ID:} Performance of AV-LLMs and OLMs on sample cases of \benchmarkname}
    \label{fig:QE_PSP_ID}
\end{figure*}

Coming to the ID task, we actually found two distinct failure modes during our evaluations. Firstly, most of the models were able to focus on the correct half of the screen for the answer. However, we found that the reasoning traces provided by the models did not contain any references to the spatial location that the model "attended to", to determine the answer, even when explicitly prompted to do so. Therefore, it is possible that the models got the answers correct due to the less challenging nature of the questions or the video data, and not their ability to spatially ground their responses. Secondly, for this particular video-question sample, the object that was asked about in the question, i.e., the \textit{"dark polka-dot shirt"}, was only shown at a close-up for the second half of the video. For the first half, the view in the video was mostly a wide-angle camera shot. From a distance, the polka-dot pattern on the shirt is not visible, and hence the shirt looks like an ordinary dark shirt. This shows that some of the models like Ola, UnifiedIO-2, Baichuan-Omni, and VideoLLaMA do not have good temporal context. Table \ref{tab:id} summarizes the model performances on the ID task.

\begin{figure*}[h!]
    \centering
    \vfill
    \includegraphics[width=\linewidth]{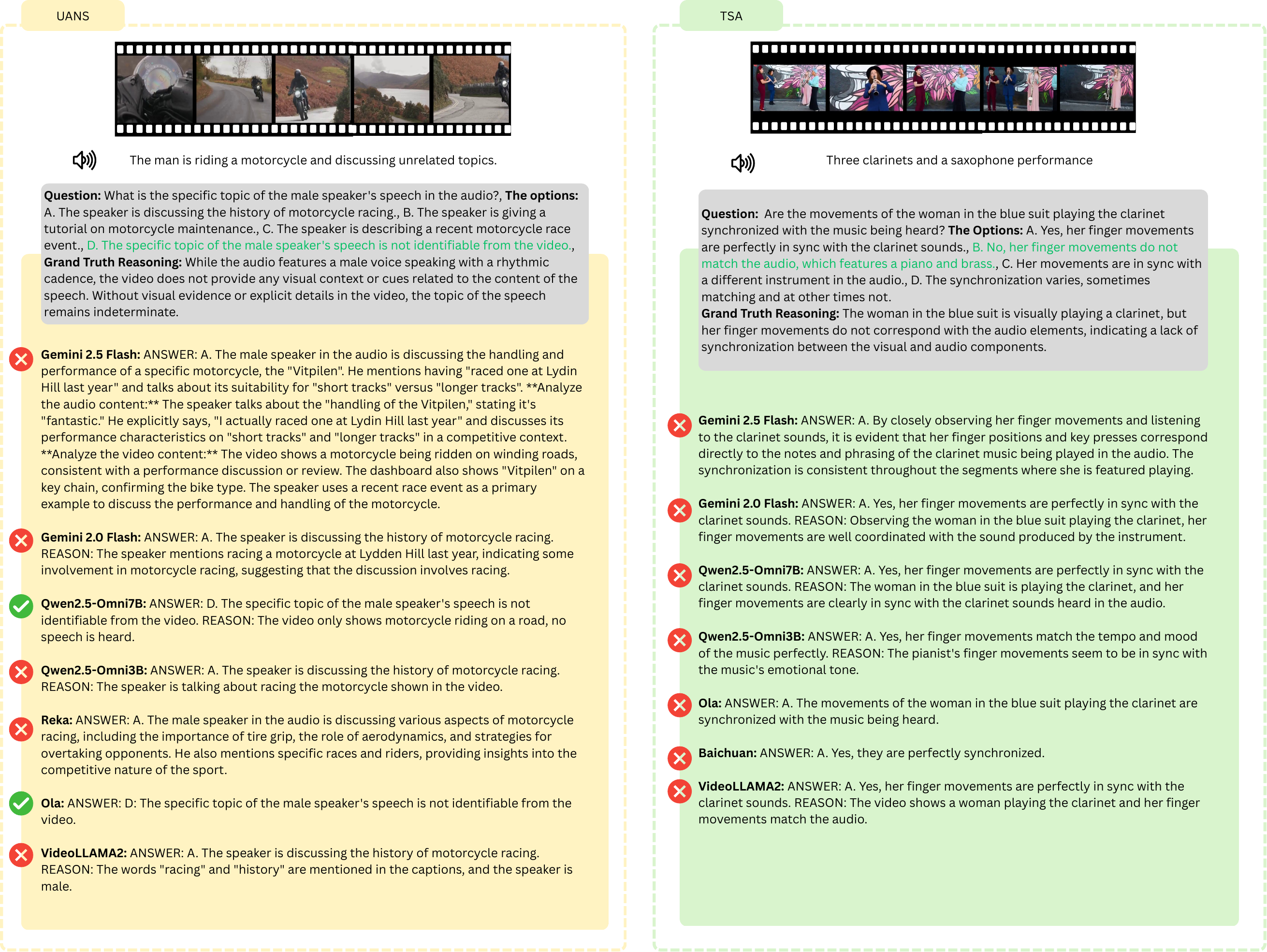}
    \caption{\textbf{UANS \& TSA:} Performance of AV-LLMs and OLMs on sample cases of \benchmarkname}
    \label{fig:QE_UANS_TSA}
    \vfill
\end{figure*}

\begin{table}[h!]
    \centering
    \begin{tabular}{lc}
    \toprule
    \textbf{Model Performance} & \textbf{Count} \\
    \midrule
    Correct Answer + Spatial Reasoning & 2 \\
    Correct Answer, No Spatial Reference & 5 \\
    Poor Temporal Context Understanding & 4 \\    
    \bottomrule
    \end{tabular}
    \caption{\textbf{ID:} Model Performance Summary}
    \label{tab:id}
\end{table}

Now in Figure \ref{fig:QE_UANS_TSA}, we have shown the several observations that we made during our evaluations on the \textit{Unanswerability} (UANS) and \textit{Tempo/AV-Synchronization Analysis} (TSA) tasks. For the UANS task, the goal was to prompt the model to answer a question about a factual detail that was not discussed in the video and verify if the model refrained from falsifying information. For this particular example, we chose a video where there is a person riding a motorcycle while talking about random, unclear topics. Upon evaluation, however, we observed that most of the models responded that the speaker was discussing the  \textit{"history of motorcycle racing"}. While the speaker does not talk about the history of anything, really, he does talk about a race. We believe that the models hallucinate a false truth, biased by the available facts. This proves the exact shortcoming of these models that we were trying to establish. While two of the models, Qwen2.5-Omni 7B and Ola, gave the correct answer, their reasoning was either absent or incorrect. A summary of the results on this task for this particular data sample can be found in Table~\ref{tab:uans}.

\begin{table}[h!]
    \centering
    \begin{tabular}{lc}
    \toprule
    \textbf{Model Performance} & \textbf{Count} \\
    \midrule
    Correct Refusal + Valid Reasoning & 0 \\
    Correct Refusal, Poor Reasoning & 2 \\
    Hallucinated False Information & 9 \\  
    \bottomrule
    \end{tabular}
    \caption{\textbf{UANS:} Model Performance Summary}
    \label{tab:uans}
\end{table}

The second task in Figure \ref{fig:QE_UANS_TSA} is TSA. In this task, we are assessing the models for their ability to recognize temporal misalignment in the audio and visual cues in the video. The data sample that we have shown here is from the misaligned category, where the audio was shifted by $\sim 5s$ to create a perceptible mismatch of audio-visual streams. The evaluation showed that almost every model misclassified the visual cues as being perfectly synchronized with the audio. This can also be supported by the extremely low accuracy scores that were observed for this task across all models. This shows that current SOTA AV-LLMs and OLMs have a very poor understanding of audio-visual tempo and synchronization. The summary of the model performances of this example case can be found in Table~\ref{tab:tsa}.

\begin{table}[h!]
    \centering
    \begin{tabular}{lc}
    \toprule
    \textbf{Model Performance} & \textbf{Count} \\
    \midrule
    Correctly Detected Misalignment & 0 \\
    Misclassified as Synchronized & 11 \\
    \bottomrule
    \end{tabular}
    \caption{\textbf{TSA:} Model Performance Summary}
    \label{tab:tsa}
\end{table}

\begin{figure*}[h!]
    \centering  
    \includegraphics[width=\textwidth]{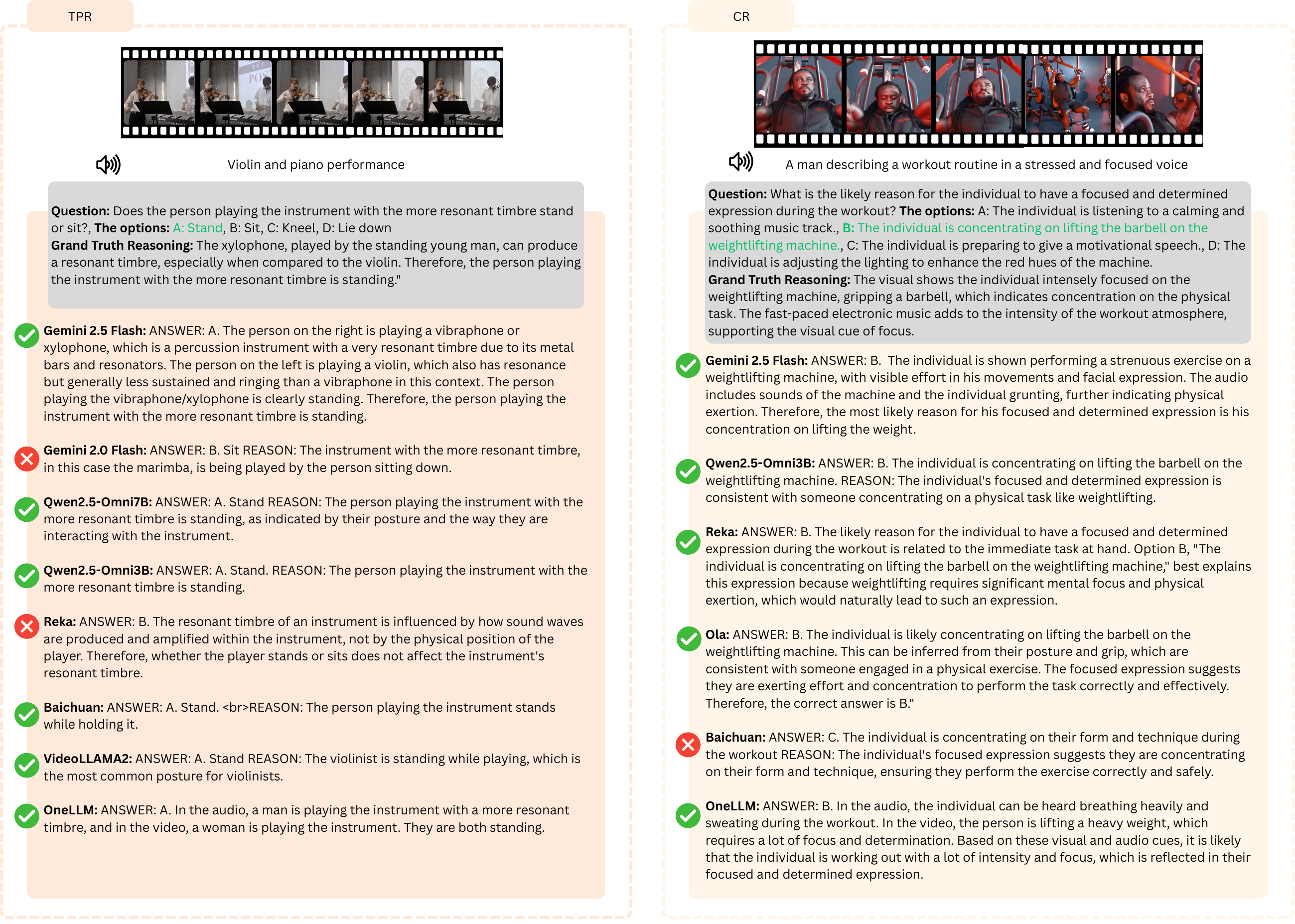}
    \caption{\textbf{TPR \& CR:} Performance of AV-LLMs and OLMs on sample cases of \benchmarkname}
    \label{fig:QE_TPR_CR}
\end{figure*}

The final two tasks are \textit{Timbre/Pitch Reasoning} and \textit{Causal Reasoning} as shown in Figure \ref{fig:QE_TPR_CR}. Starting with the TPR task, it is designed to assess a model's ability to understand and distinguish between fine-grained advanced audio attributes such as timbre \& pitch. The videos have been chosen to have two or more sound sources (mostly musical instruments), and the questions are tailored to ask the model to compare audio features among those sources. In the example shown, we have a video of two violin and xylophone players performing together, and we ask the models a question about whether the person playing the \textit{more resonant} instrument is standing or sitting. Now, in most cases, the violin is a more resonant instrument. But in this particular video, the xylophone has a more resonant timbre while the violin is barely audible. On testing, we found that every model except Gemini 2.5 Flash failed in some manner. Gemini 2.5 showed a strong perception and understanding of properties of sound/audio, such as timbre and resonance. For example, Qwen2.5-Omni models get the answer correct by saying that the performer in question is \textit{standing}, but in their reasoning, they do not clarify as to which specific performer it is and what instrument the performer is playing. This is critical, as in this particular video, both the performers are standing, so the Qwen models might be hallucinating the correct response in this case. Our suspicion is further strengthened by the fact that while VideoLLaMA-2 gets the answer correct, investigation of its reasoning traces reveals that the model was referring to the violinist while answering the question. A summary of the observations of model performances for this example can be found in Table~\ref{tab:tpr}.

\begin{table}[h!]
    \centering
    \begin{tabular}{lc}
    \toprule
    \textbf{Model Performance} & \textbf{Count} \\
    \midrule
    Correct Answer + Faithful Reasoning & 1 \\
    Correct Answer, Unfaithful Reasoning & 6 \\
    Incorrect Answer & 4 \\  
    \bottomrule
    \end{tabular}
        \caption{\textbf{TPR:} Model Performance Summary}
    \label{tab:tpr}
\end{table}

And finally, we show the observations of the CR task. This task is designed to assess a model's ability to identify the causal relationship between two events in the video. In this instance, the video shows a person working out on a seated chest-press machine. His expression is focused and determined, and the models were asked what the reason was behind this facial expression. While most of the models get the answer correct again in this case, only Gemini 2.5 Flash and OneLLM get the reasoning correct. All the other models provide unfaithful reasoning behind the correct answer. Some of the examples are the Qwen2.5-Omni model associates the individual's expressions to \textit{"someone concentrating on a physical task"}, rather than connecting it to the exercise that he is performing in the video. Similarly, Reka's reasoning says that the expression of the individual is related to the workout but does not state their causal dependency. The summary of the CR task model performance has been shown in Table~\ref{tab:cr}.

\begin{table}[h!]
    \centering
    \begin{tabular}{lc}
    \toprule
    \textbf{Model Performance} & \textbf{Count} \\
    \midrule
    Correct Answer + Causal Reasoning & 2 \\
    Correct Answer, Missing Causal Link & 8 \\
    Incorrect Answer & 1 \\
    \bottomrule
    \end{tabular}
        \caption{\textbf{CR:} Model Performance Summary}
    \label{tab:cr}
\end{table}

\section{Additional Related Works}
\subsection{Multi-modal Learning}
Multi-modal learning, when compared to traditional single modality learning \cite{vdesirr, maw}, has seen significant advancements in various domains, including cross-modal generation \cite{adverb, melfusion, magnet, vlmnav, foleygen, tang2024codi}, audio-visual representation learning \cite{listentopixel, avnav, audvisum, egoadapt, gao2024audio, sudarsanam2025representation}, multi-modal large language models \cite{aurelia, meerkat, avtrustbench, videosalmonn, videollama, zhu2023minigpt, vistallm, videollava}, and cross-modal integration \cite{li2021align, ghosh2024exploring, safari, zhang2022glipv2, zhai2023siglip, intentometer, tschannen2025siglip, volta, guo2025audio, beats, gong2025avs, egovlpv2, sun2025listen}. Recent contributions to cross-modal generation have utilized visual and/or language context to produce coherent and complex audio \cite{adverb, melfusion, tang2024codi, qwen2.5omni}. In the context of vision-language integration, recent studies illustrate the effectiveness of alignment across modalities \cite{zhuang2024falip, apollo, yu2024attention, xing2024survey}. Collectively, these initiatives highlight the necessity of dynamic, embodied perception for the development of general-purpose multi-modal systems.

\subsection{Audio-Visual Question Answering}

Recent research in open-ended audio-visual question answering (AVQA) has explored strategies for richer multi-modal reasoning, bias mitigation, and integration with large language models (LLMs). \cite{ye2024diverse} introduces a framework for answering diverse questions via text attached with key audio-visual clues, explicitly coupling linguistic queries with salient spatio-temporal cues to improve answer grounding. \cite{ma2024look} tackles bias in AVQA by rephrasing questions and restructuring training data, improving generalization across varied linguistic forms. \cite{li2024object} proposes object-aware adaptive-positivity learning, integrating object-level reasoning with adaptive confidence calibration to better handle diverse AVQA scenarios.

Bridging AVQA with multi-modal large language models, \cite{r:35} presents CAT, an enhanced MLLM architecture designed to operate in dynamic audiovisual environments, demonstrating competitive zero-shot performance on open-ended video QA datasets. \cite{chen2023autoevalvideo} introduces AutoEval-Video, an automated benchmark for assessing large vision-language models on open-ended video QA, providing a systematic and reproducible evaluation protocol. \cite{choudhury2023zero} proposes ProViQ, a procedural program-based zero-shot AVQA method that encodes reasoning steps explicitly, enabling interpretable multi-modal inference.

These works expand AVQA beyond fixed MCQ-style question-answering, emphasizing open-ended generation, bias resilience, object-aware reasoning, LLM integration, procedural reasoning, and long-form temporal understanding. These directions are complementary to \benchmarkname’s focus on structured, multi-category, and reasoning-based evaluation.



\nocite{*}                    
\bibliographystyle{plainnat}  
\bibliography{aaai2026}

\end{document}